%% file: sample-authordraft.tex
\documentclass[sigconf,authordraft,review=false]{acmart}
\input{macros.tex}
\begin{document}
\settopmatter{printacmref=false} 
\renewcommand\footnotetextcopyrightpermission[1]{} 
\pagestyle{plain} 
\title{\modelname: End-to-End Trajectory Prediction for Heterogeneous Road-Agents in Dense Traffic with Noisy Sensor Inputs}


\author{Rohan Chandra}
\affiliation{%
  \institution{University of Maryland, College Park}}

\author{Uttaran Bhattacharya}
\affiliation{%
  \institution{University of Maryland, College Park}
}

\author{Christian Roncal}
\affiliation{%
 \institution{University of Maryland, College Park}
}

\author{Aniket Bera}
\affiliation{%
  \institution{University of North Carolina, Chapel Hill}
}

\author{Dinesh Manocha}
\affiliation{%
  \institution{University of Maryland, College Park}
}
%




\input{0-Abstract.tex}
\begin{teaserfigure}
  \includegraphics[width=\textwidth]{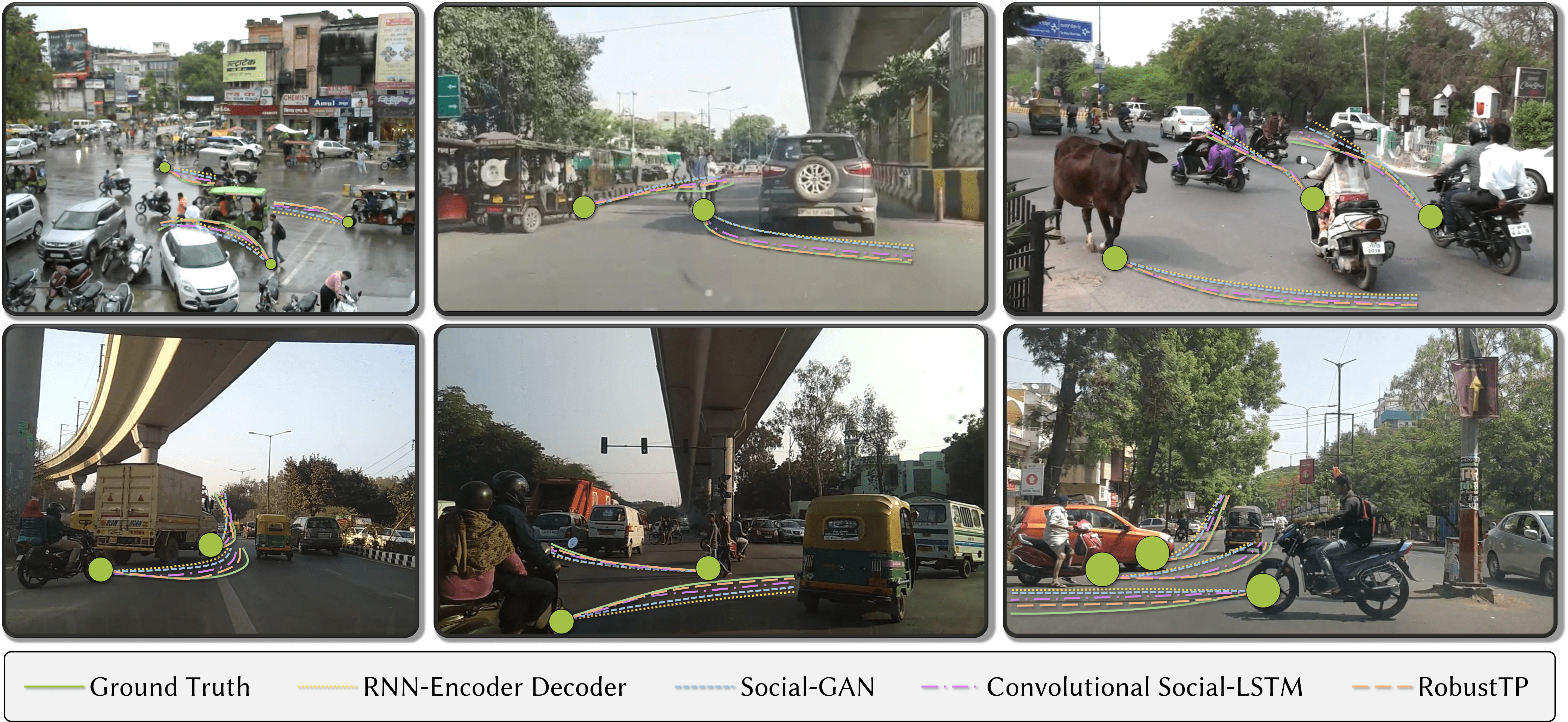}
  \caption{Trajectory Prediction Results: We highlight the performance of various end-to-end trajectory prediction methods on the TRAF dataset with different types of road-agents. We showcase six scenarios with different densities, heterogeneity, camera positions (fixed or moving), times of the day, and weather conditions. We highlight the predicted trajectories (over $5$ seconds) of some of the road-agents in each scenario to avoid clutter. The ground truth (GT) trajectory is drawn as a solid green line and prediction results for our approach, \modelname~are shown as a solid red line. The prediction results of other methods (RNN-Encoder Decoder, Social-GAN, CS-LSTM) are drawn with different dashed lines. The green circles denote the road-agent whose trajectory is being predicted. \modelname~predictions are closest to GT in all the scenarios. We observe up to 18\% improvement in average displacement error and up to 35.5\% final displacement error over prior methods for dense, heterogeneous traffic.}
  \label{fig:teaser}
\end{teaserfigure}

\maketitle

\input{1-Introduction.tex}
\input{2-RelatedWork.tex}

\input{3-TP.tex}

\input{4-RobustTP.tex}

\input{5-Results.tex}


\input{6-Conclusion.tex}

\clearpage
\bibliographystyle{acm}
\bibliography{refsnew}
\end{document}

%% file: macros.tex
\usepackage{subcaption}     
\usepackage{multirow}

\newcommand{\softwarename}{TrackNPred}
\newcommand{\modelname}{RobustTP}

\newcommand{\sota}{state-of-the-art}

\DeclareMathOperator*{\argmin}{argmin} 

\newcommand{\tim}[1]{\textrm{#1}}
\newcommand{\mc}[1]{\mathcal{#1}}
\newcommand{\bb}[1]{\mathbb{#1}}

\newcommand{\norm}[1]{\left\lVert#1\right\rVert}
\newcommand{\bcc}[1]{\left\{{#1}\right\}}
\newcommand{\brr}[1]{\left({#1}\right)}
\newcommand{\bss}[1]{\left[{#1}\right]}

%% file: 0-Abstract.tex
\begin{abstract}
We present \modelname, an end-to-end algorithm for predicting future trajectories of road-agents in dense traffic with noisy sensor input trajectories obtained from RGB cameras (either static or moving) through a tracking algorithm. In this case, we consider noise as the deviation from the ground truth trajectory. The amount of noise depends on the accuracy of the tracking algorithm. Our approach is designed for dense heterogeneous traffic, where the road agents corresponding to a mixture of buses, cars, scooters, bicycles, or pedestrians. \modelname~is an approach that first computes trajectories using a combination of a non-linear motion model and a deep learning-based instance segmentation algorithm. Next, these noisy trajectories are trained using an LSTM-CNN neural network architecture that models the interactions between road-agents in dense and heterogeneous traffic. Our trajectory prediction algorithm outperforms \sota~methods for end-to-end trajectory prediction using sensor inputs. We achieve an improvement of upto 18\% in average displacement error and an improvement of up to 35.5\% in final displacement error at the end of the prediction window (5 seconds) over the next best method. All experiments were set up on an Nvidia TiTan Xp GPU. Additionally, we release a software framework, \softwarename. The framework consists of implementations of \sota~tracking and trajectory prediction methods and tools to benchmark and evaluate them on real-world dense traffic datasets. 
\end{abstract}

%% file: 1-Introduction.tex
\section{Introduction}\label{sec:intro}

Increasingly powerful GPUs and advanced computer vision tools have made it possible to perform end-to-end, realtime tracking of heterogeneous road-agents such as cars, pedestrians, two-wheelers, etc. These tools can further be used for many applications such as autonomous driving, surveillance, action recognition, and collision-free navigation. In addition to tracking, these tools are also essential to predicting the future trajectory of each road agent. The predicted trajectories are useful for performing safe autonomous navigation, traffic forecasting, vehicle routing, and congestion management~\cite{schreier2014bayesian,DBLP:journals/corr/abs-1801-06523}.

In this paper, we focus on dense traffic composed of heterogeneous road agents. The traffic density corresponds to the number of distinct road agents captured in a single frame of the video or the number of agents per unit length (\textit{e.g.}, a kilometer) of the roadway. High-density traffic is described as traffic with more than 100 road agents per kilometer. The heterogeneity corresponds to different types of road agents with varying dynamics such as cars, buses, pedestrians, two-wheelers (scooters and motorcycles), three-wheelers (rickshaws), animals, etc. These agents have different shapes, dynamic constraints, and levels of maneuverability. The difficulty of performing trajectory prediction increases in such traffic because the trajectory of any single road-agent is affected by other road-agents in close proximity. To accurately predict the trajectory, a model for interaction with nearby road-agents needs to be considered.

Many methods have been proposed for end-to-end trajectory prediction. The agent types can be both pedestrians~\cite{social-lstm,social-gan} and traffic road-agents~\cite{nachiket,tp,traphic}. However, a major disadvantage common to all of the above methods is that they rely on manually annotated trajectories that are often not readily available for training their models. Additionally, training manually annotated trajectories introduces dataset bias and results in over-fitting, thereby generating results that do not emulate real traffic scenarios. Our goal in this paper is to perform robust end-to-end trajectory prediction using sensor trajectories. The sensors, in this case, are static or moving RGB cameras. The trajectories are obtained using a tracking algorithm and thus contains noise. In this case, we consider noise as the deviation or perturbation from the ground truth trajectory. The amount of noise depends on the accuracy of the tracking algorithm.



The chance of collisions with other road-agents in close proximity increases in such traffic. Advanced Driver Assistance Systems~(ADAS) help prevent or reduce the likelihood of traffic accidents by mitigating the adverse effects caused by human errors. ADAS collects information from a road-agent's surroundings and utilizes that information to implement critical actions to assist drivers. Predicting road-agents' trajectories is a crucial task for any ADAS to avoid collisions. Some ADAS applications for predicting a road-agent's trajectory have been proposed~\cite{adas1,adas2,adas3}. However, the methods are designed for simple road conditions with sparse traffic. Moreover, some approaches are computationally expensive as they rely on lidar-based 3D point clouds~\cite{adas3}. Some algorithms are controls-based systems that are susceptible to highly dynamic environments such as dense and heterogeneous traffic~\cite{adas2}.

\textbf{Main Contributions: }
\vspace{-5pt}
\begin{enumerate}
    \item We present an end-to-end trajectory prediction approach, \modelname, for road-agents in dense and heterogeneous traffic. The input to our algorithm is a video captured using a static or moving RGB camera and the output is the predicted trajectory over a span of $3$-$5$ seconds. We outperform \sota~methods for end-to-end trajectory prediction that use sensor inputs for training their models. We achieve an increase of up to 18\% average displacement accuracy and an increase of 35.5\% final displacement accuracy over the next best method. All experiments were set up on an Nvidia TiTan Xp GPU. Finally, \modelname~is a proof of concept for an Advanced Driver Assistance Systems~(ADAS) application and can be integrated into current ADAS applications.
    
    \item Additionally, we build a software framework, \softwarename, that contains implementations of many different tracking and trajectory prediction methods, including our novel algorithm, and tools for quickly applying them to other dense traffic datasets. The purpose of \softwarename~is twofold: to reduce collisions by computing safer paths in dense traffic and to create a common interface for many trajectory prediction approaches. \softwarename~can also benchmark different algorithms and generate performance comparisons using standard error measurement metrics on real-world dense and heterogeneous traffic datasets.
\end{enumerate}

%% file: 2-RelatedWork.tex
\section{Related Work}\label{sec:rw}

\begin{figure*}[t]
    \centering
    \includegraphics[width = \linewidth]{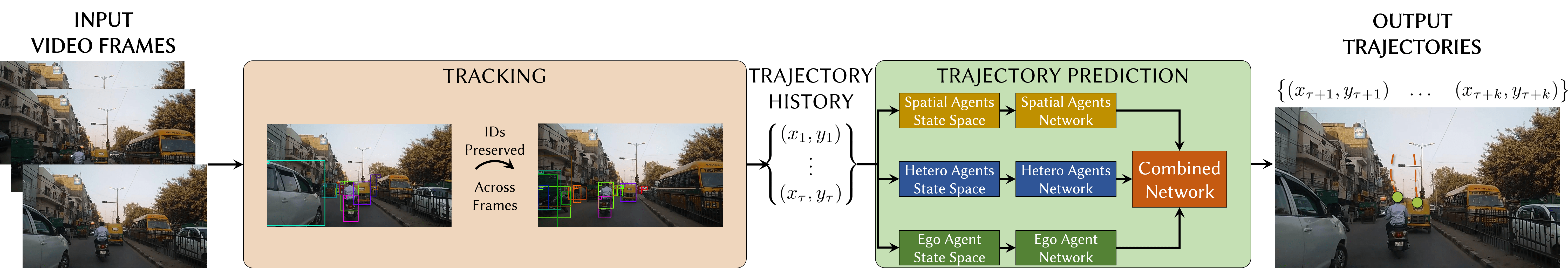}
    \caption{Overview of \modelname: \modelname~is an end-to-end trajectory prediction algorithm that uses sensor input trajectories as training data instead of manually annotated trajectories. The sensor input is an RGB video from a moving or static camera. The first step is to compute trajectories using a tracking algorithm (light orange block). The details of the tracking algorithm are provided in Section~\ref{tracking section}. The trajectories generated are the training data for the trajectory prediction algorithm (green block), the details of which are provided in Section~\ref{TP section}. The model trains on $\tau=3$ seconds of trajectory history and predicts trajectory for the next $k=5$ seconds. As an example, the predicted trajectories for two of the agents are shown in the output image at the right end. The green circles denote the positions of the agents at the beginning of prediction, as seen from a top-view in the 3D world. The red-dashed lines denote the predicted trajectories for the next 5 seconds, as seen from the same top-view in the 3D world.}
    \label{pipeline}
\end{figure*}

In this section, we give a brief overview previous work in trajectory prediction and Advanced Driver Assistance Systems~(ADAS).
\vspace{-10pt}
\subsection{Trajectory Prediction}
Trajectory prediction has been well studied through multiple approaches such as the Bayesian formulation \cite{early1-bayesian}, the Monte Carlo simulation \cite{early5}, Hidden Markov Models (HMMs) \cite{early2-HMM}, control-based systems~\cite{adas3} and kalman filters \cite{early4-kalman}. However, these methods are highly susceptible to dense and dynamics environments. \cite{adas3} is additionally computationally expensive as it relies on lidar-based 3D pointclouds.

Another line of research investigates trajectory prediction by modeling interactions between the road-agents, either explicitly or implicitly. These methods work by reducing the task of trajectory prediction to one of predicting sequences using neural networks. The sequence prediction model uses the past trajectories as training data and predicts spatial coordinates that form the future trajectory sequence. Examples of methods that explicitly model road-agent interaction include techniques based on social forces~\cite{earlysocial,early6}, velocity obstacles~\cite{van2011reciprocal}, LTA~\cite{eth}, etc. Many of these models were designed to account for interactions between pedestrians in a crowd (\textit{i.e.} homogeneous interactions) and improve the prediction accuracy~\cite{bera2016glmp}. Techniques based on velocity obstacles have been extended using kinematic constraints to model the interactions between heterogeneous road agents~\cite{autorvo}. On the other hand, there are some methods~\cite{traphic} that model the heterogeneous interactions between road agents implicitly.

Approaches based on deep learning use Recurrent Neural Networks (RNNs) and its variants for sequence modeling. 
The benefits of RNNs for sequence modeling makes them a reasonable choice for traffic prediction. Since RNNs cannot utilize information from too far back in its memory, many traffic trajectory prediction methods use long short-term memory networks (LSTMs) to predict trajectory sequences. These include algorithms to predict trajectories in traffic scenarios with few heterogeneous interactions~\cite{nachiket,tp}. These techniques have also been used for trajectory prediction for pedestrians in a crowd~\cite{social-lstm, vermula2018social}.

RNN-based methods can be combined with other deep learning architectures to form hybrid networks for trajectory prediction. Some examples of deep learning architectures include CNNs, GANs, VAEs, and LSTMs. Each architecture has its own set of advantages and, for many tasks, the accuracy of the performance can be increased by combining the advantages of individual architectures. For example, generative models have been successfully used for tasks such as super resolution~\cite{hybridsuper}, image-to-image translation~\cite{hybridit}, and image synthesis~\cite{hybriddraw}. However, their application in trajectory prediction has been limited because back-propagation during training is non-trivial. In spite of this, generative models such as VAEs and GANs have been used for trajectory prediction of pedestrians in a crowd~\cite{social-gan} and in sparse traffic \cite{lee2017desire}. Alternatively, Convolutional Neural Networks (CNNs or ConvNets) have also been successfully used in many computer vision applications like object recognition \cite{objrecogreview}. Recently, they have also been used for traffic trajectory prediction \cite{cnnpredict1,cnnpredict2}.

\subsection{Advanced Driver Assistance Systems~(ADAS)}

Passive safety measures (that do not process sensory information) in vehicles include safety belts, brakes, airbags etc. ADAS are active safety measures that collect and process sensory information through sensors such as lidars, radars, stereo cameras, and RGB cameras. Various ADAS process the input information in different ways to implement actions that assist the driver and prevent or reduce the likelihood of traffic accidents due to human error. The development of ADAS began with the Anti-Lock Braking System (ABS) introduced into production in the late 1970s.

ADAS for trajectory prediction have been proposed~\cite{adas1,adas2,adas3}. However, \cite{adas1} uses a constant acceleration motion model to predict trajectories, which is unrealistic in dense and heterogeneous traffic. \cite{adas2} uses a Model Predictive Control based system, which is susceptible to highly dynamic environments. Lastly, \cite{adas3} uses a kalman filter approach and relies on lidars to collect 3D pointcloud information which is computationally expensive. 

As ADAS with various functionality become popular, it is not uncommon for multiple systems to be installed on a vehicle. If each function uses its own sensors and processing unit, it will make installation difficult and raise the cost of the vehicle. As a countermeasure, research integrating multiple functions into a single system has been pursued and is expected to make installation easier, decrease power consumption, and vehicle pricing. \modelname~contributes towards this research effort by integrating realtime tracking with trajectory prediction.

In addition to to trajectory prediction applications, several other interesting ADAS are currently being used in vehicles on the road. For example, the Adaptive Cruise Control (ACC) automatically adapts speed to maintain a safe distance from vehicles in front. The Blind Spot Detection (BSD) helps drivers when they pull out in order to overtake another road-agent. Emergency Brake Assist (EBA) ensures optimum braking by detecting critical traffic situations. When EBA detects an impending collision, the braking system is put on emergency standby. Intelligent Headlamp Control (IHC) provides optimal night vision. The headlamps are set to provide optimum lighting via a continuous change of the high and low beams of the lights.

%% file: 3-TP.tex
\section{\modelname: Overview and Algorithm}\label{sec:traphic}
In this section, we begin by formally stating the problem and describing the notation. Then we give an overview of our approach to realtime end-to-end trajectory prediction in dense and heterogeneous traffic scenarios. 
\begin{figure*}[t]
    \centering
    \includegraphics[width = \linewidth]{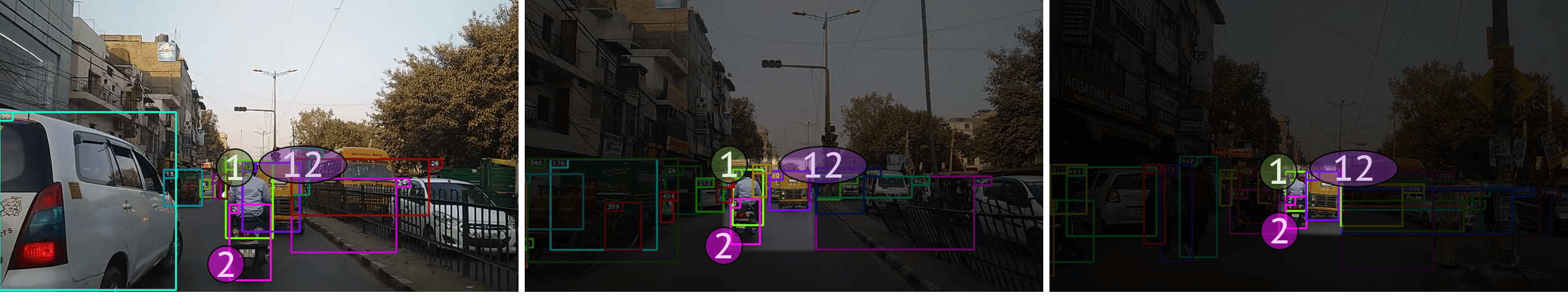}
    \caption{Qualitative analysis of our tracking algorithm on the TRAF dataset consisting of approximately 30 road-agents in the traffic scene. Frames are chosen with a gap equal to the length of the prediction window, \textit{i.e.} 5 seconds($\sim$ 150 frames). Each bounding box color is associated with a unique ID (displayed on the top left corner of each bounding box). \textbf{Observation:} Note the consistencies in the (ID)(color); for example, observe rider+two-wheeler (1-2)(green-pink) and three-wheeler (12)(purple).}
\end{figure*}
\label{tracking results}

\subsection{Problem Setup and Notation}

Given a set of $N$ road agents $\mathcal{R}=\{r_i\}_{i=1\dots N}$, the trajectory history of each road agent $r_i$ over $\tau$ frames, denoted $\mathcal{T}_i = \{(x_1,y_1), (x_2,y_2),$ $\ldots,(x_\tau,y_\tau)\}$, and the road agent's size $l$, we predict the trajectory, \textit{i.e.}, the spatial coordinates of that road agent for the next $k$ frames. 

We define the state space of each road agent $r_i$ as
\begin{equation}
    \Omega_i := \begin{bmatrix}\mathcal{T}_i & \Delta\mathcal{T}_i & c & l\end{bmatrix}^\top,
\end{equation}
where $\Delta$ is a derivative operator that is used to compute the velocity of the road agent, and $c:=[c(x_1,y_1),\dots,c(x_\tau,y_\tau)]^\top$. The traffic concentration, $c(x,y)$, at the location $(x,y)$, is defined as the number of road agents between $(x,y)$ and $(x,y) + (\delta x, \delta y)$ for some predefined $(\delta x, \delta y)>0$.

We also compute camera parameters from given videos using standard techniques~\cite{cameracalib1, cameracalib2} and use the parameters to estimate the camera homography matrices. The homography matrices are subsequently used to convert the location of road agents in 2D pixels to 3D world coordinates w.r.t. a predetermined frame of reference, similar to approaches in \cite{social-gan,social-lstm}. All state-space representations are subsequently converted to the 3D world space.

Finally, we consider a method to be more robust compared to other methods if the trajectories predicted by it are less affected by noise in the trajectory history (arising due to sensor artifacts, inaccuracies in tracking and similar factors).

\subsection{Tracking by Detection}
\label{tracking section}
Manually labeled training data are not representative of real-world trajectories and thus do not produce realistic results. Moreover, they are not easily available for dense and heterogeneous traffic scenarios.

\begin{figure}[t]
    \centering
    \includegraphics[width=\columnwidth]{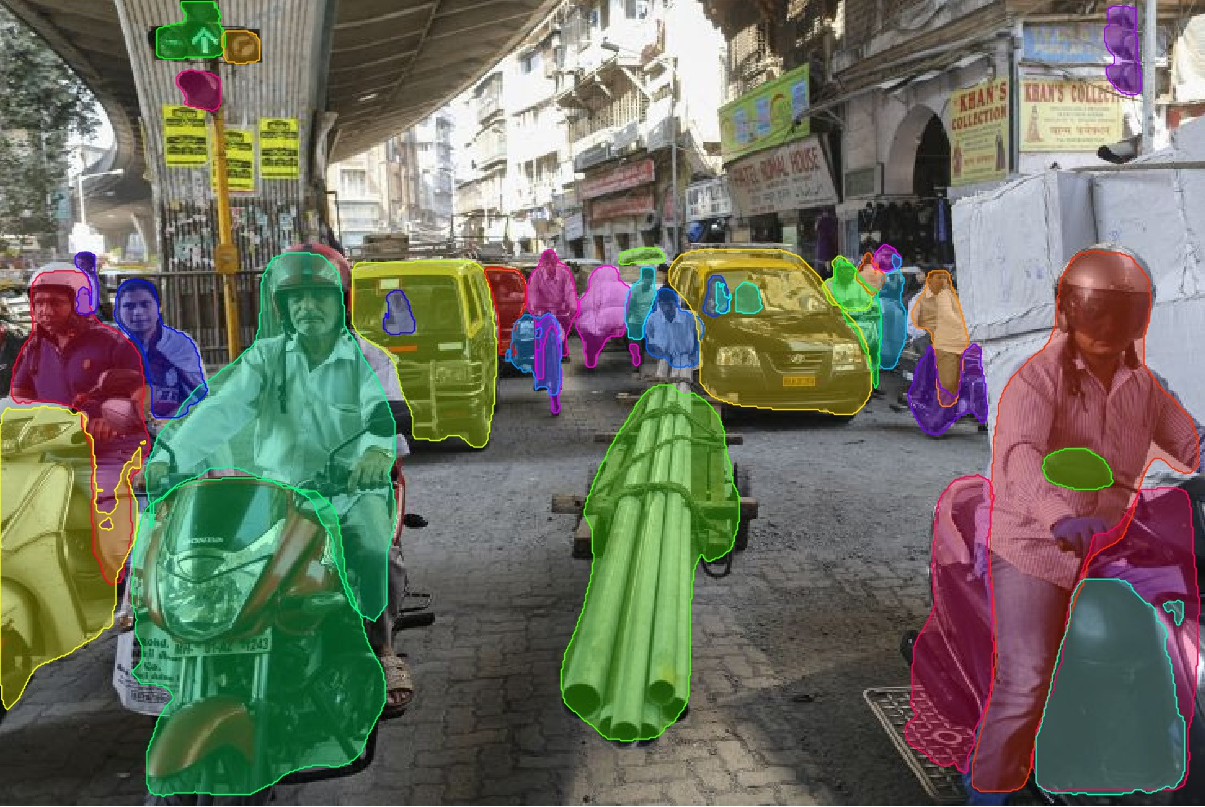}
    \caption{Efficient representations of road-agents in dense traffic. Instance segmentation helps reduce occlusions by reducing background clutter otherwise present in typical bounding box representations. These representations are used in our tracking by detection algorithm. The different colors simply denote individual road-agents.}
    \label{fig:unpred_masked}
    \vspace{-15pt}
\end{figure}

Instead, we implement an end-to-end system where a tracking algorithm computes the input trajectories that form the training set for the trajectory prediction algorithm. We first use instance segmentation based on the Mask R-CNN algorithm to generate background-subtracted representations of the identified road-agent at some time step $t$. Then, our tracking methodology uses the RVO collision avoidance formulation to predict the next state, $\Omega$, for a road-agent at time $t+1$. Finally, the algorithm computes the ID of the road-agent at time $t+1$ using a feature matching process described in~\cite{deepsort}. We now describe the tracking and detection processes separately in detail:

\textbf{Detection: }The main difficulty in detecting road-agents in dense traffic from a front camera can be attributed to an increased likelihood of occlusions. In such instances, bounding boxes are an inefficient visual representation. Therefore, we use Mask R-CNN to perform instance segmentation to reduce background clutter and generate efficient and occlusion-free representations (Figures~\ref{fig:unpred_masked}).


Mask R-CNN begins by generating a set of bounding boxes for each road-agent in each frame. $\mc{B}= \{ \bb{B} \ | \ \bb{B} = [\langle x,y \rangle_{\textrm{top left}}, w, h, s, r]$, $r_i \in \mc{H} \}$ denotes the set of bounding boxes for each $r_i$ at current time $t$, where $\langle x,y \rangle_{\textrm{top left}}, w, h, s, \tim{and} \ r$ denote the top left corner, width, height, scale, and aspect ratio of $\bb{B}$, respectively.

Mask R-CNN also outputs a set of masks, $\mc{M}$, for each corresponding bounding box. That is, $\mc{M} = \{\bb{M} \ | \ r_i \in \mc{H}\}$ denotes the set of masks for each $r_i$ at current time $t$, where each $\bb{M}$ is a $[w \times h]$ matrix of boolean variables. Now, let $\mc{W}= \{\bb{W}(\cdot) \ | \ r_i \in \mc{H} \}$ be the set of white canvases where each canvas, $\bb{W} = [1]_{w \times h}$, $w$ and $h$ are the width and height of each $\bb{B}$ at current time $t$. Then, 

    \[\mc{S} = \{ \bb{W}(\bb{M}) \ | \ \bb{W} \in \mc{W}, \bb{M} \in \mc{M}, r_i \in \mc{H} \},\]
    
\noindent is the set of background-subtracted segmented representations for each road-agent in the current frame. Note that during each iteration of the detection process, the IDs of the road-agents are known, and the task is to identify the road-agents at the next time-step.

\textbf{Tracking: } Our tracking by detection algorithm uses a non-linear motion model to predict the next state of the agent. Prior motion models with constant velocity or constant acceleration assumptions have been shown to not accurately model dense scenarios as they do not take into account collision avoidance behavior. Reciprocal Velocity Obstacles (RVO)~\cite{van2011reciprocal} models collision avoidance behavior for dense scenes. RVO can be applied to pedestrians in a crowd as well as dense traffic environments. We use the RVO formulation and modify the formulation for segmented road-agents.

\noindent The RVO formulation requires an additional parameter $v_\textrm{pref}$. $v_{\tim{pref}}$ is the velocity the pedestrian would have taken in the absence of obstacles or colliding pedestrians, computed using the standard RVO formulation.

The computation of the new state, $\Omega_{t+1}$, is formulated as an optimization problem. For each road-agent, RVO computes a feasible region where it can move without collision. This region is defined according to the RVO collision avoidance constraints (or ORCA constraints~\cite{van2011reciprocal}). If the ORCA constraints forbid a road-agent's preferred velocity, that road-agent chooses the velocity  closest to its preferred velocity that lies in the feasible region, as given by the following optimization:
\begin{equation}
    v_{\textrm{new}} = \argmin_{\substack{v \notin ORCA}} ||v - v_{\textrm{pref}}||,
\end{equation}
The velocity, $v_{\textrm{new}}$, is then used to calculate the new position of a traffic agent.

To combine instance segmentation with RVO, we modify the state vector, $\Omega_{t}$, to include bounding box information by setting the position to the centers of the bounding boxes. The centers of the bounding boxes are by extension, centers of each segmented road-agent.

\subsection{Trajectory Prediction}
\label{TP section}
The goal is to predict trajectories,  \textit{i.e.} temporal sequences of spatial coordinates of a road agent using neural networks. In standard neural networks, the objective function can be stated as,

\begin{equation}
\min_w \norm{y - \phi \brr{w^Tx}},
\label{nneq}
\end{equation}

\noindent where $x$ is the input trajectory of a road-agent, $\phi$ is the non-linearity, and $\phi \brr{w^Tx}$ is the predicted trajectory of the road-agent. The objective is to learn a set of weights, $w$, that minimizes~\ref{nneq}. However, the difficulty of trajectory prediction in dense traffic lies in the fact that a road-agent's trajectory is affected by the trajectories of other road-agents in close proximity, not just it's own. This is especially prominent when agents are heterogeneous. For example, the maneuverability of a bus differs significantly from that of pedestrians; a pedestrian can change directions quickly while a bus cannot. Additionally, behavioral cues should also be emphasized in heterogeneous traffic. Aggressive drivers have non-uniform trajectories due to maneuvers like over-speeding, tailgating, and overtaking, while conservative agents tend not to stray from their current trajectories. However, the ability of non-linearity functions in neural networks to model behavioral aspects of human drivers is still an unsolved problem. This leads to a need for a model that accounts for interactions of a road-agent with other nearby road-agents.

In \cite{traphic}, the authors propose such a model that considers two forms of interactions: spatial interactions based on the location of a neighboring road-agent with respect to the ego road-agent and heterogeneous interactions that consider the different static parameters (size) and dynamic parameters (steering angle) of road-agents inside a pre-computed neighborhood. The spatial interactions are motivated by the observation that, in dense traffic, a road-agent focuses primarily on road-agents that are in its horizon view (this is defined in the original paper). In heterogeneous interactions, however, the road-agent learns to assign adaptive weights to different heterogeneous road-agents. Combining the two interactions, the objective function~\ref{nneq} is replaced by,

\begin{equation}
    \min_w \norm{y - \bcc{\phi \brr{w_x^Tx} +\phi \brr{w_s^Ts}+ \phi \brr{w_h^Th}  }},
\end{equation}

\noindent where $w = \bss{w_x~w_s~w_h}^T$ is a vector comprising the weights for the input for the ego road-agent, the road-agents modeled by the spatial interaction formulation, and the road-agents modeled by the heterogeneous interaction formulation, respectively. Temporal sequence prediction requires neural networks that can capture temporal dependencies in data, such as LSTMs~\cite{graves}. However, LSTMs cannot learn dependencies or relationships of various heterogeneous road agents because the parameters of each individual LSTM are independent of one another. In this regard, ConvNets have been used in computer vision applications with greater success because they can learn locally dependent features from images. Thus, to leverage the benefits of both, the authors of~\cite{traphic} combine ConvNets with LSTMs to learn locally useful relationships (both in space and in time) between heterogeneous road agents.

%% file: 4-RobustTP.tex
\section{TrackNPred: A Software Framework for End-to-End Trajectory Prediction}\label{sec:arch}

\begin{figure}[t]
    \centering
    \includegraphics[width = \columnwidth]{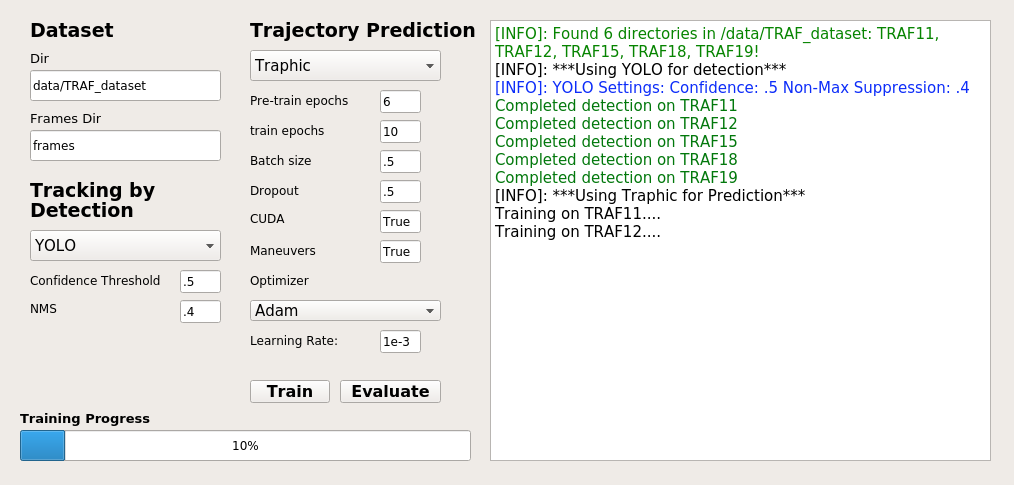}
    \caption{\softwarename~is a deep learning-based framework that integrates trajectory prediction methods with tracking by detection algorithms to motivate further research in end-to-end trajectory prediction. In this figure, we show the graphical user interface of \softwarename~where one can select the tracking by detection algorithm as well as choose the trajectory prediction method. The user can also set the hyperparameters for the training and evaluation phases. If the input can be connected to an RGB camera mounted on a road-agent, then \softwarename~can be extended to ADAS applications.}
    \label{gui}
    \vspace{-15pt}
\end{figure}

\softwarename~is a python-based software library~\footnote{https://tinyurl.com/RobustTP} for end-to-end realtime trajectory prediction for autonomous road-agents. 
Our first goal, through \softwarename, is to enable autonomous road-agents to navigate safely in dense and heterogeneous traffic by estimating how road-agents, that are in close proximity, are going to move in the next few seconds. 

The continuous advancement in deep learning has resulted in the development of several \sota~ tracking and trajectory prediction algorithms that have shown impressive results on real world dense and heterogeneous traffic datasets. However, there are currently no theoretical guarantees to validate the comparison of performance of different deep learning models. It is only through empirical research that one can evaluate the efficiency of a particular deep learning model. 

Our second goal is to equip researchers with a packaged deep learning tool that performs trajectory prediction based on various \sota~neural network architectures, such as Generative Adversarial Networks (GANs~\cite{social-gan}), Recurrent Neural Networks (LSTMs~\cite{social-lstm, traphic}, and Convolutional Neural Networks (CNNs~\cite{nachiket,traphic}). Therefore, one of the advantages of \softwarename~is that it enables researchers to experiment with these different deep learning architectures with minimal difficulty. Researchers need only select hyperparameters for the chosen network. We also provide the ability to modify individual architectures without disrupting the rest of the methods (Figure~\ref{gui}).

\softwarename~integrates realtime tracking algorithms with end-to-end trajectory prediction methods to create a robust framework. The input is simply a video (through a moving or static RGB camera). \softwarename~selects a tracking method from the tracking module to first generate a trajectory, $\mathcal{T}_i = \{(x_1,y_1), (x_2,y_2),$ $\ldots, (x_n,y_n)\}$, for the $i^\textrm{th}$ road-agent for $n$ frames, where $n$ is a constant. The trajectories for each agent are then treated as the trajectory history for that agent in the trajectory prediction module. The final output is the future trajectory for the ego-agent, $\mathcal{T}_\textrm{ego} = ((x_{n+1},y_{n+1}), (x_{n+2},$ $y_{n+2}), \ldots, (x_{n+k},y_{n+k}))$, where $k$ is the length of the prediction window. This is a major difference from trajectory prediction methods in the literature~\cite{nachiket,social-gan,social-lstm,tp,traphic} that rely on manually annotated input trajectories. \softwarename, in contrast, does not require any ground truth trajectories.

Finally, \softwarename~evaluates and benchmarks realtime performances of various trajectory prediction methods on a real-world traffic dataset\footnote{https://go.umd.edu/TRAF-Dataset}~\cite{traphic}. This dataset contains more than $50$ videos of dense and heterogeneous traffic. The dataset consists of the following road agent categories: cars, buses, trucks, rickshaws, pedestrians, scooters, motorcycles, and other road agents such as carts and  animals. Overall, the dataset contains approximately $13$ motorized vehicles, $5$ pedestrians, and $2$ bicycles per frame. Annotations consist of spatial coordinates, an agent ID, and an agent type. The dataset is categorized according to camera viewpoint (front-facing/top-view), motion (moving/static), time of day (day/evening/night), and density level (sparse/moderate/heavy/\newline challenging). All the videos have a resolution of $1280\times 720$.

\subsection{Methods Implemented in \softwarename}
One of our goals is to motivate research in highly accurate, end-to-end, and realtime trajectory prediction methods. To achieve this goal, we design a common interface for several state-of-the-art methods from both tracking and trajectory prediction literature. Such a design facilitates easy bench-marking of new algorithms with respect to the \sota. The methods in \softwarename~differ in numerous ways from their original implementations in the literature in order to achieve improved accuracy in tracking and prediction in dense and heterogeneous traffic. Table~\ref{tab: tracknpred algo list} provides a list of algorithms currently implemented in \softwarename.

\input{tracknpred.tex}

\textbf{Tracking Module: }For tracking, we mainly focus our attention on tracking by detection approaches. These are approaches that leverage deep learning-based object detection models. This is because tracking methods that do not perform detection require manual, near-optimal initialization of each road-agent's state information in the first video frame. 
Further, methods that do not utilize object detection need to know the number of road-agents in each frame a priori so they do not handle cases in which new road-agents enter the scene during the video. Tracking by detection approaches overcome these limitations by employing a detection framework to recognize road-agents entering at any point during the video and initialize their state-space information.

At present, we implement python-based tracking by detection algorithms to facilitate easy integration into \softwarename. DeepSORT~\cite{deepsort} is currently the \sota~realtime tracker implemented in python. Naturally, we use DeepSORT as the base tracker. However, DeepSORT was originally developed using a constant velocity model with the goal of tracking pedestrians in sparse crowds. Consequently, it is not optimized for dense and heterogeneous traffic scenes that may contain cars, buses, pedestrians, two-wheelers, and even animals. Therefore, we replace the constant velocity model with a non-linear RVO motion model~\cite{van2008reciprocal}, which is designed for motion planning in dense environments. 

The advantage of using tracking by detection algorithms is that we can combine the unique benefits of different object detection models. For example, we integrate two \sota~object detection models, YOLO and Mask R-CNN. They are \sota~in its own category. The YOLO algorithm is extremely fast as compared to Mask R-CNN wile the latter offers a higher accuracy.

The output of the tracking module is a trajectory file with corresponding ID's. An ID is an integer unique to every agent. Each row of this file corresponds to the following format:

\[<\textrm{Fid}>,<\textrm{Vid}>,<\textrm{center-X}>,<\textrm{center-Y}>\]

\noindent which denotes the frame ID, vehicle ID, and the 2D coordinates of the center of the bounding box of the road-agent. This trajectory file is input for the trajectory prediction module.
\newline

\textbf{Trajectory Prediction Module: }\softwarename~currently supports the following end-to-end trajectory prediction algorithms: Social-Gan~\cite{social-gan}, Convolutional Social-LSTM~\cite{nachiket}, RNN Encoder-Decoder~\cite{rnn_ed}, and TraPHic~\cite{traphic}. All trajectory prediction methods that are implemented in \softwarename~work in essentially the same manner. However, there are some differences which we highlight. Social-GAN~\cite{social-gan} was originally trained to predict the trajectories of pedestrians in a crowd. Additionally, CS-LSTM~\cite{nachiket} was designed to predict trajectories for road-agents in sparse and homogeneous traffic. Our goal is to perform trajectory prediction in dense and heterogeneous traffic environments. Therefore, we trained all three implementations on a real world dense and heterogeneous traffic dataset.

%% file: tracknpred.tex
\begin{table}[t]
    \centering
    \caption{The list of algorithms currently implemented in \softwarename.}
    \resizebox{\columnwidth}{!}{
    \label{tab: tracknpred algo list}
    \begin{tabular}{|c|c|}
        \hline
        & Methods\\
        \hline
        \multirow{2}{*}{Tracking by Detection} & Mask R-CNN + DeepSORT \\
        & YOLO + DeepSORT \\
        \hline
        \multirow{4}{*}{Trajectory Prediction} & RNN- Encoder Decoder~\cite{rnn_ed} \\
        & Social-GAN~\cite{social-gan} \\
        & Covolutional Social-LSTM~\cite{nachiket}\\
        & TraPHic~\cite{traphic} \\
        \hline
    \end{tabular}
    }
    \vspace{-15pt}
\end{table}

%% file: 5-Results.tex
\section{Experimental Evaluation}
\label{sec:results}
We first compare the performance of our trajectory prediction algorithm, \modelname, with trajectory prediction methods trained on manually annotated trajectories, on the TRAF dataset~\cite{traphic} in Section~\ref{sec:noisytp}. We also compare with the same methods when all are trained on sensor inputs. We use our software framework, \softwarename~for the second set of experiments in Section~\ref{sec:softlib}.

\subsection{Comparison with Methods Using Manually Annotated Inputs}
\label{sec:noisytp}
 We use the following well-known metrics for evaluation:
\begin{itemize}
    \item Average Displacement Error (ADE): The average of the root mean squared error (RMSE) between the ground truth and the predicted trajectory position at every time frame for the entire duration of 5 seconds. A lower ADE for a method implies that the method has a lower drift from the ground truth on the average, which is desirable.
    \item Final Displacement Error (FDE): The RMSE between the ground truth and the predicted trajectory position at the last time frame. A lower FDE for a method indicates it has a better prediction in the longer term.
\end{itemize}
\input{compare_methods.tex}
\input{robusttp.tex}
In Table~\ref{tab:compare_methods}, we compare our algorithm that is trained on sensor inputs (using Mask R-CNN~\cite{maskrcnn} for detection), with existing methods that, by contrast, use manually annotated ground truth trajectory history as training data. The RNN Encoder-Decoder model~\cite{rnn_ed} is used for sequence modeling in many applications and has been adapted for trajectory prediction~\cite{social-lstm,nachiket}. However, these approaches mainly target either sparse traffic or sparse pedestrian crowds. Therefore, we trained it from scratch on the TRAF dataset. Social-GAN~\cite{social-gan} is state-of-the-art for predicting trajectories of pedestrians in sparse scenarios, we, therefore, trained this method on the TRAF dataset from scratch as well. The CS-LSTM~\cite{nachiket}, is designed for trajectory prediction for road-agents for sparse traffic, therefore we fine-tuned on top of it with the TRAF dataset.

After preparing the methods for fair evaluation, we observe that even with noisy trajectories as input, the ADE of our algorithm is 1.49 meters lower than the RNN Encoder-Decoder model~\cite{rnn_ed} and 1.98 meters lower than Social-GAN~\cite{social-gan}. This is primarily because Social-GAN~\cite{social-gan} is trained for a single road-agent (pedestrians) in sparse scenarios, which does not transfer to modeling heterogeneous agents in dense traffic in the TRAF dataset. Additionally, our FDE is 2.72 meters and 2.35 meters lower than RNN-ED and S-GAN, respectively.

\begin{figure*}[t]
    \centering
    \subcaptionbox{Using Mask R-CNN~\cite{maskrcnn} for detection.\label{fig:rmse_mrcnn}}%
    [0.45\linewidth]{\includegraphics[width=0.45\linewidth]{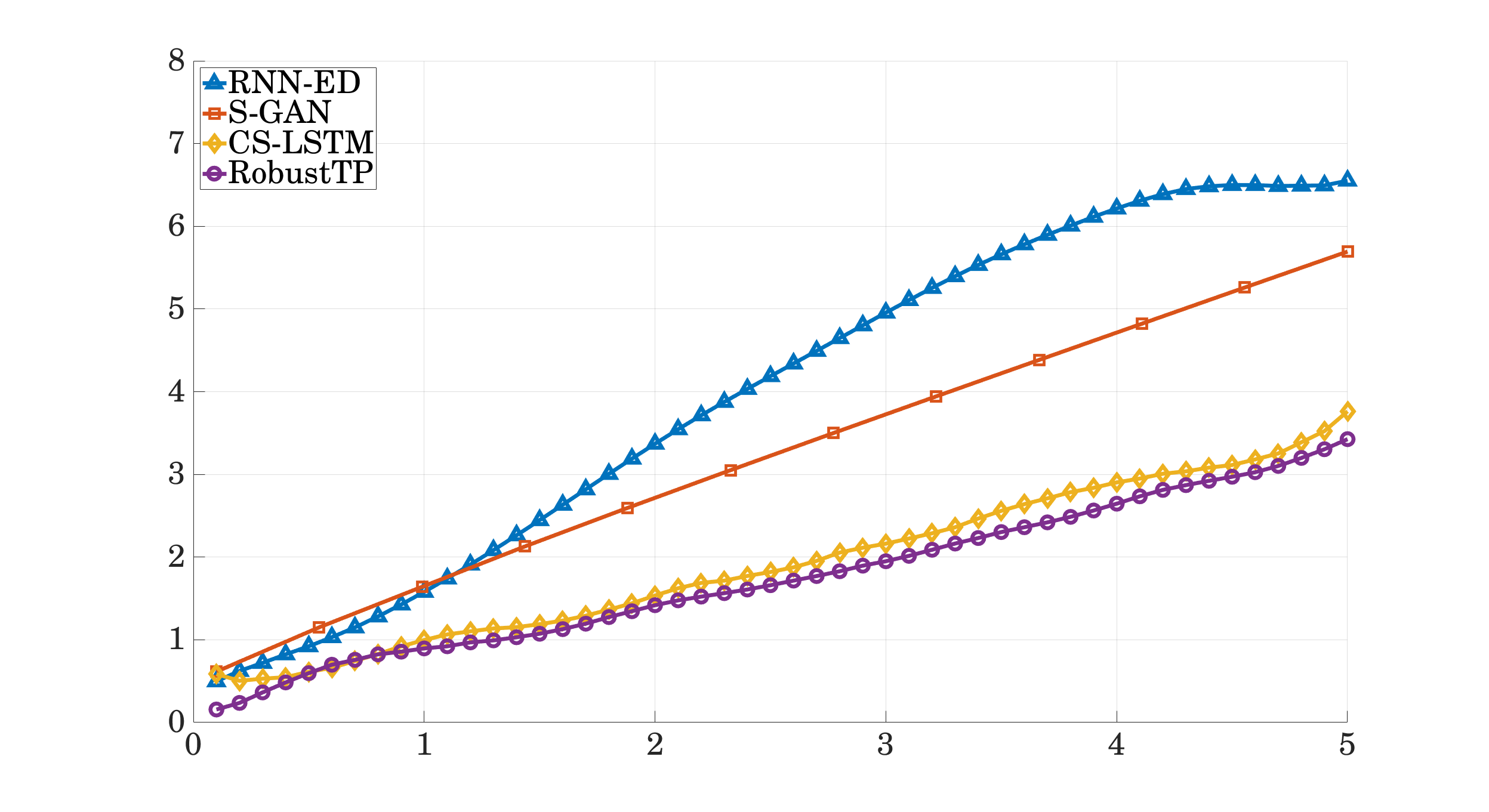}}\hfill
    \subcaptionbox{Using YOLO~\cite{yolo} for detection.\label{fig:rmse_yolo}}%
    [0.45\linewidth]{\includegraphics[width=0.45\linewidth]{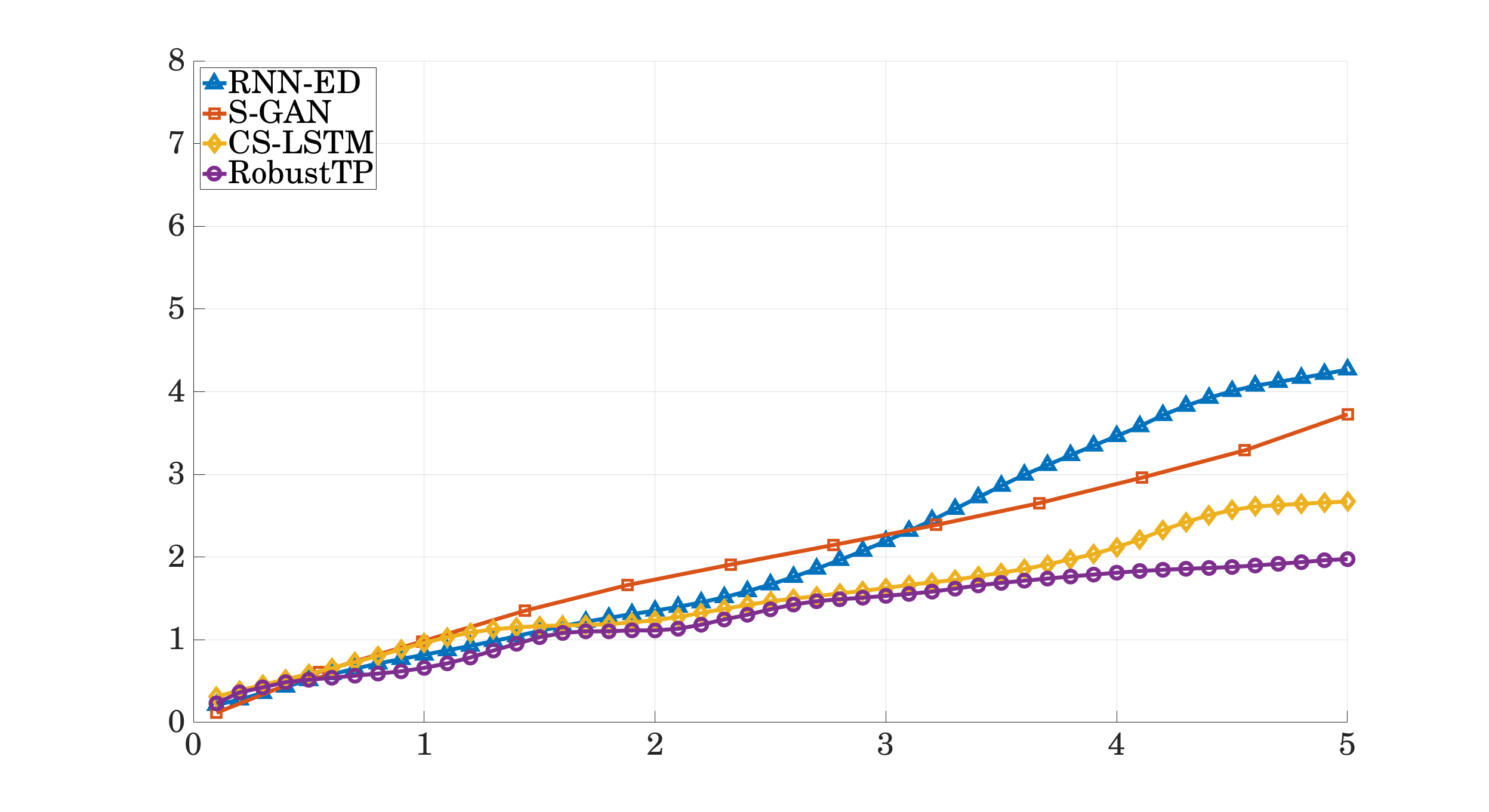}}
    \caption{\textbf{RMSE Curve Plot: }We compare the RMSE-s of \modelname~with \sota~end-to-end trajectory prediction methods on the TRAF dataset. All methods were trained using sensor inputs instead of manually annotated trajectories. The graphs were generated using \softwarename. Overall, we observe that using YOLO for detection provides lower RMSE-s for all the methods. More importantly, the RMSE of \modelname~is consistently lower than all the other methods after the first second of prediction, implying it has better longer-term performance.}
    \label{fig:rmse}
\end{figure*}

 We also observe that the ADE of our algorithm is 0.6 meters higher than CS-LSTM~\cite{nachiket}. CS-LSTM weighs all agents in the neighborhood of the ego-agent equally. Thus it cannot adapt to dense, heterogeneous traffic. However, this problem is offset by the fact that we used manually annotated trajectory history as input to the CS-LSTM, resulting in overall better performance. Finally, compared to TraPHic~\cite{traphic}, our ADE is only slightly (0.97 meters) higher, since TraPHic also models dense and heterogeneous road-agents and uses ground truth trajectory history for prediction. Overall upon comparison, we note that \modelname~is at par with the state-of-the-art performance of TraPHic and CS-LSTM, while being able to achieve high accuracy by training on trajectories generated from sensor inputs.

\subsection{Comparison with Methods Using Sensor Inputs}\label{sec:softlib}
We use our software framework, \softwarename, to run a set of experiments to evaluate the performance of end-to-end trajectory prediction methods using sensor inputs instead of manually annotated trajectories on the TRAF dataset~\cite{traphic}. We use the output of two tracking methods in the library --- one uses Mask R-CNN~\cite{maskrcnn} for detection and the other uses YOLO~\cite{yolo} for detection --- to generate the input trajectory histories. We then use each of these trajectory histories as input to each of the trajectory prediction methods to obtain the trajectory predictions for the next 5 seconds. Finally, we compare and contrast their performances in Table~\ref{tab:robusttp}. We also compare the RMSE curves produced by all these methods in Figure~\ref{fig:rmse}. For each detection model, we can observe that, when all the methods use trajectory history from noisy tracking data, our method has a clear advantage. For example, in the case of Mask R-CNN~\cite{maskrcnn} as the detection model, in addition to outperforming both RNN-ED~\cite{rnn_ed} and Social-GAN~\cite{social-gan} by 2.24 meters and 1.48 meters respectively, \modelname~also outperforms CS-LSTM by 0.16 meters on the ADE metric. It is also \sota~on the FDE metric. The RMSE curves in Figure~\ref{fig:rmse} further show that beyond the first second of prediction, the RMSE of \modelname~is consistently lower than all the other methods. Moreover, at the end of 5 seconds, the final RMSE of \modelname~is well below 4 meters (less than the length of an average car). Thus, \modelname~is more reliable than the other methods in longer-term prediction, which is a crucial benefit to consider when deciding trajectory prediction methods for real-world applications.

Finally, these experiments also serve to highlight how \softwarename~makes it convenient to benchmark the tracking and trajectory prediction methods that best suits specific end-to-end trajectory prediction tasks, thereby encouraging further research on novel end-to-end trajectory prediction algorithms.


%% file: compare_methods.tex
\begin{table}[t]
    \centering
    \caption{We evaluate \modelname~with methods that use manually annotated trajectory histories, on the TRAF Dataset. The results are reported in the following format: ADE/FDE, where ADE is the average displacement RMSE over the 5 seconds of prediction and FDE is the final displacement RMSE at the end of 5 seconds. We observe that \modelname~is at par with the \sota. }
    \resizebox{\columnwidth}{!}{
    \label{tab:compare_methods}
    \begin{tabular}{|c|c|c|c|c|}
        \hline
        RNN-ED & S-GAN & CS-LSTM & TraPHic & \modelname \\
        \hline
        3.24/5.16 & 2.76/4.79 & 1.15/3.35 & \textbf{0.78/2.44} & 1.75/3.42 \\
        \hline
    \end{tabular}
    }
\end{table}

%% file: robusttp.tex
\begin{table}[t]
    \centering
    \caption{We evaluate \modelname~with methods that use noisy sensor input, on the TRAF Dataset. The trajectory histories are computed using tracking by two detection methods: Mask R-CNN~\cite{maskrcnn} and YOLO~\cite{yolo}. The results are reported in the following format: ADE/FDE, where ADE is the average displacement RMSE over the $k$ seconds of prediction and FDE is the final displacement RMSE at the end of $k$ seconds. We tested for both short-term ($k=3$) and longer-term ($k=5$) predictions. We observe for all the cases that \modelname~is the \sota. }
    \resizebox{\columnwidth}{!}{
    \label{tab:robusttp}
    \begin{tabular}{|r|c|c|c|c|}
        \hline
        \multicolumn{5}{|c|}{Prediction length, $k=3$ secs} \\
        \hline
        & RNN-ED & S-GAN & CS-LSTM & \modelname \\
        \hline
        MRCNN & 2.60/4.96 & 2.11/3.50 & 1.27/2.01 & \textbf{1.14/1.90} \\
        \hline
        YOLO & 1.13/2.18 & 1.29/2.18 & 1.08/1.55 & \textbf{0.96/1.53} \\
        \hline
        \multicolumn{5}{|c|}{Prediction length, $k=5$ secs} \\
        \hline
        & RNN-ED & S-GAN & CS-LSTM & \modelname \\
        \hline
        MRCNN & 3.99/6.55 & 3.23/5.69 & 1.91/3.76 & \textbf{1.75/3.42} \\
        \hline
        YOLO & 2.06/4.26 & 1.98/3.72 & 1.52/2.67 & \textbf{1.29/1.97} \\
        \hline
    \end{tabular}
    }
\end{table}

%% file: 6-Conclusion.tex
\section{Conclusion, Limitations, and Future Work}\label{sec:conclusion}
We presented a novel end-to-end algorithm, RobustTP, for predicting the trajectories of road agents in dense and heterogeneous traffic. Our approach does not require manually annotated trajectories for training our model. We use 3 seconds of trajectory history as input and predict the next 5 seconds of the road-agent's trajectory. 

\modelname~has some limitations. The size of the TRAF dataset prohibits training larger deep learning networks. Therefore, we cannot guarantee generalization to all forms of dense and heterogeneous traffic scenarios.

Regarding future work, \modelname~is a proof of concept and can be designed as an effective ADAS. The resulting ADAS would improve upon existing \sota~trajectory prediction ADAS in many ways. First it is applicable to dense and heterogeneous traffic. Prior trajectory prediction-based ADAS~\cite{adas1,adas2,adas3} are either computationally expensive or susceptible to dynamic environments. \modelname~is computationally cost-effective since it uses a single RGB camera and the learning algorithm generalizes across scenes with varying dynamics. Second, it improves generalization to real-world scenarios. In autonomous driving, the degree to which a pre-trained trajectory prediction model generalizes the real world dynamics is a major concern. Online Hard Example Mining (OHEM) is an online machine learning training procedure wherein negatively classified training examples identified by the model would be added to the training set, and re-trained in an online manner. This has shown to increase generalization to new unseen test data, an advantage that is desired in autonomous driving. \modelname, with its end-to-end capability integrating real-world trajectories in realtime, can support OHEM and offer better generalization in dense and heterogeneous traffic. We also plan to expand \softwarename~by adding more algorithms and providing more hyperparameter tuning capabilities.